\documentclass[10pt,twocolumn,letterpaper]{article}

\usepackage{iccv}
\usepackage{times}
\usepackage{epsfig}
\usepackage{graphicx}
\usepackage{amsmath}
\usepackage{amssymb}

\usepackage[pagebackref=true,breaklinks=true,letterpaper=true,colorlinks,bookmarks=false]{hyperref}

\usepackage{booktabs}

\usepackage{graphicx}

\usepackage{bm}
\usepackage{algorithm}
\usepackage{algorithmic}
\usepackage{multirow}
\usepackage{color}

\usepackage{graphicx}
\usepackage{float} 
\usepackage{subcaption}

\iccvfinalcopy 

\def\iccvPaperID{****} 

\ificcvfinal\fi

\begin{document}

\title{RMP-Loss: Regularizing Membrane Potential Distribution for Spiking Neural Networks}

\author{Yufei Guo\thanks{Equal contribution.}, Xiaode Liu\footnote[1]{}, Yuanpei Chen, Liwen Zhang, Weihang Peng, Yuhan Zhang,\\
Xuhui Huang, Zhe Ma\thanks{Corresponding author.}\\
 Intelligent Science \& Technology Academy of CASIC, China\\
 Scientific Research Laboratory of Aerospace Intelligent Systems and Technology, China\\
{\tt\small yfguo@pku.edu.cn, lxde@pku.edu.cn, mazhe\_thu@163.com}
}

\maketitle
\ificcvfinal\thispagestyle{empty}\fi

\begin{abstract}
Spiking Neural Networks (SNNs) as one of the biology-inspired models have received much attention recently. It can significantly reduce energy consumption since they quantize the real-valued membrane potentials to 0/1 spikes to transmit information thus the multiplications of activations and weights can be replaced by additions when implemented on hardware. However, this quantization mechanism will inevitably introduce quantization error, thus causing catastrophic information loss. To address the quantization error problem, we propose a regularizing membrane potential loss (RMP-Loss) to adjust the distribution which is directly related to quantization error to a range close to the spikes.  Our method is extremely simple to implement and straightforward to train an SNN. Furthermore, it is shown to consistently outperform previous state-of-the-art methods over different network architectures and datasets.
\end{abstract}

\section{Introduction}
\label{sec:intro}

Recently, many efforts have been done to make deep neural networks (DNNs) lightweight, so that they can be deployed in devices where energy consumption is limited. To this end, several approaches have been proposed, including network pruning \cite{2017Channel}, network quantization \cite{gong2019differentiable,li2021brecq,eshraghian2022navigating},  knowledge transfer/distillation \cite{2018Model}, neural architecture search \cite{Neural2016,liu2018darts}, and spiking neural networks (SNNs) \cite{guo2023joint,guo2023neuroclip,guo2023direct,2020Incorporating,2018Direct,li2021free,ororbia2019spiking,xu2023constructing,xu2023biologically,zhou2023spikformer,shen2023esl,xu2022hierarchical,xu2021robust}. The SNN provides a special way to reduce energy consumption following the working mechanism of the brain neuron. Its neurons accumulate spikes from previous neurons and present spikes to posterior neurons when the membrane potential exceeds the firing threshold. This information transmission paradigm will convert the computationally expensive multiplication to computationally convenient additions thus making SNNs energy-efficient when implemented on hardware. Specialized neuromorphic hardware based on an event-driven processing paradigm is currently under various stages of development, \textit{e.g.}, SpiNNaker \cite{2008SpiNNaker}, TrueNorth \cite{2015TrueNorth}, Tianjic \cite{2019Towards}, and Loihi \cite{2018Loihi}, where SNNs can be efficiently implemented further. Due to the advantage of computational efficiency and rapid development of neuromorphic hardware, the SNN has gained more and more attention.

\begin{figure*}[t]
	\centering
	\includegraphics[width=1.0\textwidth]{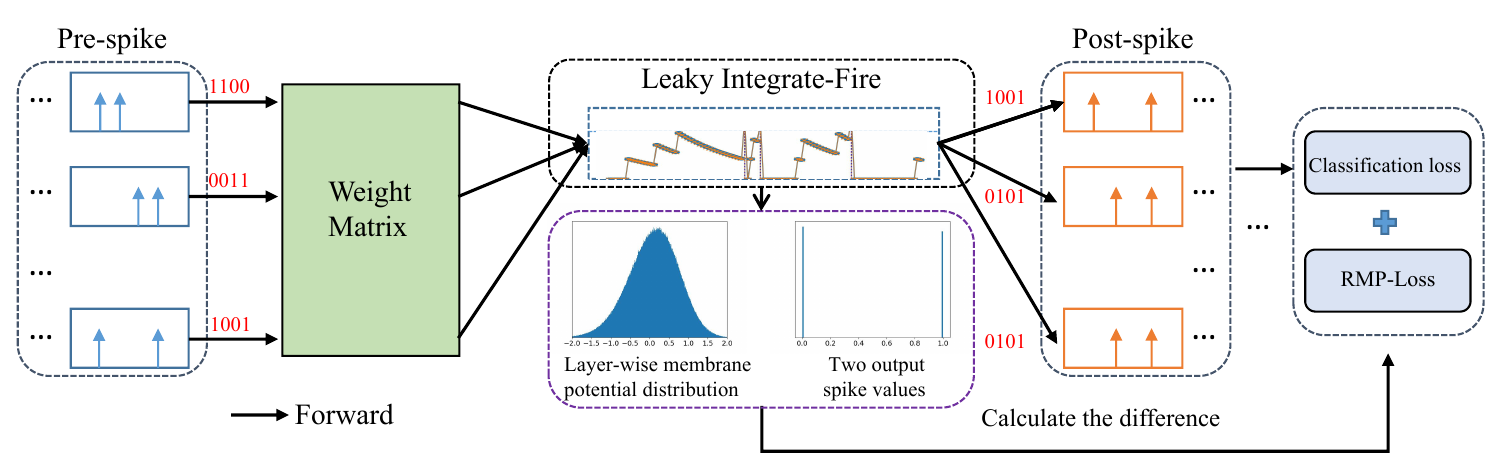} 
	\caption{The overall workflow of the proposed method. We embed a membrane potential regularization loss in the task loss to redistribute the membrane potential in the training phase to reduce the quantization error.}
	\label{workflow}
\end{figure*}

Although SNNs have been widely studied, their performance is still not comparable with that of DNNs. This performance gap is largely related to the quantization of the real-valued membrane potential to 0/1 spikes for the firing of the SNN in implementation~\cite{guo2022reducing}. The excessive information loss induced by the firing activity forcing all information only to two values will cause accuracy to decrease. Although information loss is important for computer vision tasks and the quantization will cause information loss too, the essential role of the activation function is to introduce non-linearity for neural networks~\cite{2018Activation}. Therefore, how to effectively reduce the information loss of membrane potential quantization is of high research importance. However, as far as we know, few studies have focused on directly solving this problem. 
The quantization error problem also exists in these methods that convert the ANN model to an SNN model~\cite{Deng2021Optimal,bu2022optimal,li2021free}. However, these methods solve the problem by changing the activation function in ANNs or increasing timesteps in SNNs, which don't work for SNN training. 
InfLoR-SNN~\cite{guo2022reducing} adds a membrane potential redistribution function in the spiking neuron  to reduce the quantization error via redistributing the membrane potential. However, it will decrease the biological plausibility of the SNN and increase the inference burden.
This paper focuses on reducing the quantization error in SNN training directly and aims to introduce no burden for the SNN.

Quantization error is smaller when the membrane potential is close to the spiking threshold or reset values~\cite{guo2022reducing}. Hence, to mitigate the information loss, we suggest redistributing the membrane potential to where the membrane potential is closer to the 0/1 spike. Then an additional loss term aims at \textbf{r}egularizing \textbf{m}embrane \textbf{p}otential is presented, called RMP-Loss, which can encourage the membrane potentials to gather around binary spike values during the training phase. The workflow of our method is shown in Fig.~\ref{workflow}. Our main contributions can be concluded as follows:

\begin{itemize}
	
\item To our best knowledge, there have been few works noticing the quantization error in direct training of SNNs. To mitigate the quantization error, we present the RMP-Loss, which is of benefit to training an SNN model that enjoys a narrow gap between the membrane potential and its corresponding 0/1 spike. Furthermore, we also provide theoretical proof to clarify why the RMP-Loss can prevent information loss.

\item Some existing methods can address information loss too. While achieving comparable performance, more parameters or computation burdens are also introduced in the inference phase. Different from those methods, the RMP-Loss can handle information loss directly without introducing any additional parameters or inference burden.

\item Extensive experiments on both static and dynamic datasets show that our method performs better than many state-of-the-art SNN models.

\end{itemize}


\section{Related Work}

There are two main problems to train an SNN model~\cite{guo2023direct}. The first is that the firing activity is non-differentiable thus the SNN model cannot be trained with these back-propagation methods directly. One way to mitigate the optimization difficulty is through surrogate functions. This kind of method replaces the non-differentiable firing activity with a differentiable surrogate function to calculate the gradient in the back-propagation~\cite{2016Training,2017Spatio,2017Event,2018SLAYER}. In~\cite{2011Error}, the derivative of a truncated quadratic function was used to approximate the derivative of the firing activity function. In \cite{2017SuperSpike} and \cite{2020LISNN}, the derivatives of a sigmoid \cite{2017SuperSpike} and a rectangular function were respectively adopted to construct the surrogate gradient. Furthermore, a dynamic evolutionary surrogate gradient that could maintain accurate gradients during the training was proposed in \cite{li2021differentiable,guoloss,chedifferentiable,ChenGradual2022}. Another way to solve the non-differentiable problem is converting an ANN model to an SNN model, known as ANN-SNN conversion methods~\cite{2020RMP,2020Deep,li2021free,2019Spiking,bu2022optimized,hao2023bridging,hao2023reducing}.
The ANN-SNN method trains a high-accuracy ANN with ReLU activation first and then transforms the network parameters into an SNN under the supposition that the activation of the ANN can be approximated by the average firing rates of an SNN. However, conversion methods will introduce conversion errors inevitably. Many efforts were made to tackle this problem, such as long inference time~\cite{2015Fastclassifying}, threshold rescaling~\cite{2019Going}, soft reset~\cite{2020RMP}, threshold shift~\cite{li2021free}, and the quantization clip-floor-shift activation function~\cite{bu2022optimal}. This paper adopts the surrogate gradient method.

The second problem is information loss. Though transmitting information with binary spikes is more efficient than that with real values, quantizing these real-valued membrane potentials to 0/1 spikes will cause information loss. There have been some works handling this problem indirectly. Such as learning the appropriate membrane leak-and-firing threshold~\cite{2020DIET, 2019Technical}. These are of benefit to finding a preferable quantization choice. Similarly, in~\cite{2020Incorporating, 2020Effective}, some learnable neuron parameters are incorporated into the SNN to optimize the quantization choice. In~\cite{Guo_2022_CVPR}, three regularization losses are introduced to penalize three undesired shifts of membrane potential distribution to make 
easy optimization and convergence for the SNN. The regularization is also beneficial to reduce quantization error. However, all these methods do not handle the information loss problem directly and most of them require more parameters. 
InfLoR-SNN~\cite{guo2022reducing} suggests adding a membrane potential redistribution function in the spiking neuron to redistribute the membrane potential closer to spike values, which can be seen as a direct method to reduce the quantization error. However, adding another function in spiking neurons will decrease the biological plausibility of the SNN and increase the inference burden.
In this work, we focus on handling the quantization error problem directly without introducing extra parameters or computation burden in the inference phase.

\section{Preliminary}
In this section, a brief review of the primary computing element of SNNs is first provided in Sec.~3.1. Then we introduce the proposed surrogate gradient that handles the non-differentiability challenge in Sec.~3.2. Finally, in Sec.~3.3, we describe the threshold-dependent batch normalization technique used in this work.

\subsection{Spiking Neuron Model}

\textit{Spiking neuron}. The primary computing element, \textit{i.e.}, the neuron of an SNN is much different from that of an ANN. The neuron of an ANN plays the role of nonlinear transformation and can output real-valued values. Meanwhile, the neuron in an SNN accumulates the information from previous neurons into its membrane potential and presents the spike to the following neurons when its membrane potential exceeds the firing threshold. The SNN is more efficient with this special information transmission paradigm while suffering the information loss problem. A unified form of this spiking neuron model can be formulated as follows,
\begin{equation}\label{ht11}
  u^{(t),{\rm pre}} = \tau u^{(t-1)} + x^{(t)},
\end{equation}
\begin{equation}\label{ot}
	o^{(t)}=
	\left\{
	\begin{array}{lll}
		1, \ \ {\rm if} \; u^{(t),{\rm pre}} \ge V_{\rm th} \\
		0, \ \ {\rm otherwise}
	\end{array},
	\right.
\end{equation}
\begin{equation}\label{vt}
	u^{(t)} = u^{(t),{\rm pre}} (1-o^{(t)}).
\end{equation}
Where $u^{(t),{\rm pre}}$ and $u^{(t)}$ are pre-membrane potential and membrane potential, respectively, $x^{(t)}$ is input information, $o^{(t)}$ is output spike at the timestep $t$. The range of constant leaky factor $\tau$ is $(0, 1)$, and we set $\tau$ as $0.25$. $V_{\rm th}$ is the firing threshold and is set to $0.5$ here.

\textit{Output layer neuron}. In the general classification task, the output of the last layer will be presented to the \texttt{Softmax} function first and then find the winner class. We make the neuron model in the output layer only accumulate the incoming inputs without any leakage as final output like doing in recent practices\cite{2020DIET,2020Going}, described by
\begin{equation}\label{lyn}
  u^{(t)} =  u^{(t-1)} + x^{(t)}.
\end{equation}
Then the cross-entropy loss is calculated according to the final membrane potential at the last timesteps,  $u^{(T)}$.

\subsection{Surrogate Gradients of SNNs}

As shown in Eq.~\ref{ot}. the firing activity of SNNs can be seen as a step function. Its derivative is $0$ everywhere except at $V_{\rm th}$. This non-differentiability problem of firing activity will cause gradient vanishing or explosion and make the back-propagation unsuitable for training SNNs, directly. As mentioned before, many prior works adopted the surrogate function to replace the firing function to obtain a suitable gradient~\cite{2020LISNN,2020Going,2018Direct}. 
In InfLoR-SNN~\cite{guo2022reducing}, an extra membrane potential redistribution function is added before the firing function. It argues that this redistribution function will reduce the information loss in forward propagation. However, investigating it from a backward propagation perspective and considering the redistribution function being next to the firing function, it can be seen as a more suitable surrogate gradient if we gather the STE (used in~\cite{guo2022reducing}) gradients of the firing function and redistribution function together. In this sense, the suitable surrogate gradient can also be seen as a route to reduce quantization error.
Therefore, in this paper, we adopt the gathered redistribution function in InfLoR-SNN as our surrogate function, which indeed performs better than other methods~\cite{2020LISNN,2020Going,2018Direct} in our experiments, denoted as
\begin{equation}\label{sg}
\varphi (u) =
\left\{
     \begin{array}{ll}
     0, & u \textless 0,  \\
     \frac{1}{2 {\rm tanh}(3/2)} {\rm tanh}(3(u-1/2))+1/2, & 0\leq u\leq 1,\\
     1, & u \textgreater 1.
     \end{array}
\right.
\end{equation}
\subsection{Threshold-dependent Batch Normalization}

Normalization techniques can effectively reduce the training time and alleviate the gradient vanishing or explosion problem for training DNNs. A most widely used technique called batch normalization (BN)~\cite{2015BN} uses the distribution of the summed input to a neuron over a mini-batch of training cases to compute a mean and variance which are then used to normalize the summed input to that neuron on each training case. This is significantly effective for convolutional neural networks (CNNs). However, directly applying BN to SNNs will ignore the temporal characteristic and can not achieve the desired effect. To this end, a more suitable normalization technique for SNNs, named threshold-dependent BN (tdBN) was further proposed in \cite{2020Going}. It normalizes feature inputs on both temporal and spatial dimensions. Besides, as its name implies, tdBN makes normalized variance dependent on the firing threshold, \textit{i.e.}, the pre-activations are normalized to $N(0,(\alpha V_{\rm th})^2)$ instead of $N(0, 1)$. this can balance pre-synaptic input and threshold to maintain a reasonable firing rate. In this paper,  we also adopt the tdBN, as follows,
\begin{equation}\label{tdbn1}
	\bar{\textbf{X}} = \alpha V_{\rm th} \frac{\textbf{X}- \mathrm{mean}(\mathbf{X})}{\sqrt{\mathrm{mean}((\mathbf{X}-\mathrm{mean}(\mathbf{X}))^2)+\epsilon}},
\end{equation}
\begin{equation}\label{tdbn2}
	\tilde{\textbf{X}} = \lambda \bar{\textbf{X}} + \beta.
\end{equation}
where $\textbf{X} \in \mathbb{R}^{T\times B\times C\times H\times W}$ ($T$: timesteps; $B$: batch size; $C$: channel; $(H, W)$: spatial domain) is a 5D tensor and $\alpha$ is a hyper-parameter set as 1.

\section{RMP-Loss}

As aforementioned, we suggest reducing the quantization error to avoid information loss in supervised training-based SNNs. In this section, we first try to provide a metric to measure the quantization error in SNNs. Then based on the metric, we introduce our method, a loss term for regularizing membrane potential, \textit{i.e.}, RMP-Loss, which can push the membrane potentials close to spike values to reduce the quantization error.  Finally, theoretical proof to clarify why the RMP-Loss can prevent information loss is given.

\subsection{Quantization Error Estimation}

Estimating the quantization error precisely is important for designing a method to reduce the prediction error. Hence, before giving the full details of the RMP-Loss, here we try to formulate the quantization error first. As shown in Eq.~\ref{ot}, the output of a neuron is dependent on the magnitude of the membrane potential. When the membrane potential exceeds the firing threshold, the output is $1$, otherwise, is $0$; that is, these full-precision values will be mapped to only two spike values. The difference between output, $o$, and membrane potential, $u$, can be measured as
\begin{equation}\label{dis}
	\mathrm{Dis}(o,u)=
	\left\{
	\begin{array}{lll}
		1 - u, \ \ {\rm if} \; u \ge V_{\rm th} \\
		0 - u, \ \ {\rm otherwise}
	\end{array}.
	\right.
\end{equation}
Obviously, the closer a membrane potential is to its corresponding spike, the smaller the quantization error. The quantization error depends on the margin between the membrane potential and its corresponding spike. Then we define the quantization error as
\begin{equation}\label{qe}
	q(u) = |\mathrm{Dis}(o,u)|^p,
\end{equation}
where $p>0$, which is usually set to $1$ and/or $2$ corresponding to L1-norm and/or L2-norm. Unless otherwise specified, we set it as 2 in the paper.

\subsection{Regularizing Membrane Potential Distribution}

To mitigate the information loss problem in SNNs, the quantization error should be reduced to some extent. To this end, we propose RMP-Loss to carve the distribution to a range close to the spikes based on $q(u)$, as
\begin{equation}\label{lrmp}
	\mathcal{L}_{\rm RMP} = \frac{1}{TLBCWH}\sum_{t,l,b,c,w,h}{q(u)}.
\end{equation}
where $T$, $L$, $B$, $C$, $W$, and $H$ denote the number of timesteps, the number of layers, batch size, channel length, width, and height of the membrane potential map, respectively. Finally, taking classification loss into consideration, the total loss can be written as
\begin{equation}\label{total}
	\mathcal{L}_{\rm CE-RMP} = \mathcal{L}_{\rm CE}+\lambda(n) \mathcal{L}_{\rm RMP},
\end{equation}
where $\mathcal{L}_{CE}$ is the cross-entropy loss, and $\lambda(n)$ is a balanced coefficient, which changes dynamically with the training epoch. Obviously, $\mathcal{L}_{RMP}$ does not take into account the classification problem at hand and will add a new constraint in the network optimization. To make the network parameter update have enough freedom at the beginning, which is important to train a well-performance network, and still focus on the classification task at the end. We adopt a strategy of increasing first and decreasing later to adjust the $\lambda(n)$ as follows,
\begin{equation}\label{lambda}
	\lambda(n)=
	\left\{
	\begin{array}{lll}
		2k \frac{n}{N}, \ \ {\rm if} \; n \le N/2 \\
		2k(1-\frac{n}{N}), \ \ {\rm otherwise}
	\end{array},
	\right.
\end{equation}
where $n$ is the $n$-th epoch and $k$ is a coefficient, which controls the amplitude of regularization. Unless otherwise specified, we set it as 0.1 in the paper.

\begin{algorithm}[tb]
	\caption{Training an SNN with RMP-Loss.}
	\label{alg:iresg}
	\textbf{Input}: An SNN to be trained; Training dataset, total number of training epochs $N$, total number of training iterations per epoch $I$.\\
	\textbf{Output}: The trained SNN.

	\begin{algorithmic}[1] 
		\FOR {All $n = 1, 2, \dots, N-th$ epoch}
		\FOR {All $i = 1, 2, \dots, I-th$ iteration}
		\STATE
		
		\textbf{Feed-Forward}:
		
		\STATE Calculate the SNN output, $\mathbf{O}^i$, membrane potential, $\mathbf{U}$, and class label, $\mathbf{Y}^i$;
		\STATE Compute classification loss $\mathcal{L}_{\rm CE}= {{\mathcal{L}}_{\rm CE}}(\mathbf{O},\mathbf{Y}^i)$;
		\STATE Compute RMP-Loss ${\mathcal{L}}_{\rm RMP} = {\mathcal{L}}_{\rm RMP}(\mathbf{U})$;
		\STATE Compute the total loss $\mathcal{L}_{\rm CE-RMP}=\mathcal{L}_{\rm CE}+\lambda(n) \mathcal{L}_{\rm RMP}$.
		
	    \textbf{Backward-Propagation}:
	    
	    \STATE Calculate the gradients $\Delta \mathbf{W} = \frac{\partial {\mathcal{L}_{\rm CE-RMP}}}{\partial \mathbf{W}} = \sum_t{\frac{\partial ({\mathcal{L}_{\rm CE}+\lambda(n) \mathcal{L}_{\rm RMP}})}{\partial \mathbf{y}^t} \frac{\partial \mathbf{y}^t}{\partial \mathbf{{U}}} \frac{\partial \mathbf{U}}{\partial \mathbf{W}}}$, where $\mathbf{y}^t$ is feature maps at $t$-th timesteps and $ \frac{\partial \mathbf{y}^t}{\partial \mathbf{{U}}}$ is the gradient of firing function, which can be calculated as $\frac{\partial \varphi (\mathbf U)}{\partial \mathbf{{U}}}$.
	    
	    \textbf{Parameters Update}:
		\STATE Update all parameters with the learning rate $\eta$.
		
		\ENDFOR
		\ENDFOR
		\STATE \textbf{return} the trained SNN.
	\end{algorithmic}
\end{algorithm}

The final resulting RMP-Loss training method is defined in Algo. 1. 

\subsection{ Analysis and Discussion}

In this work, we assume that RMP-Loss can help reduce the quantization loss of the SNNs. 
To verify our assumption, a theoretical analysis is conducted by using the information entropy concept.
To mitigate the information loss, the output spike tensor $\mathbf O$ should reflect the information of the membrane potential tensor $\mathbf U$ as much as possible.
Since KL-divergence is a metric to measure the difference between two random variable distributions. 
We use KL-divergence to compute the difference between $\mathbf O$ and $\mathbf U$, which refers to the difficulty of reconstructing $\mathbf U$ from $\mathbf O$, \textit{i.e.}, information loss degree. Then we provide the theoretical analysis below.

Let $x^{u}\sim \mathcal{U}$ where $x^{u}$ are samples $\in \mathbf{U}$ and $\mathcal{U}$ represents the distribution of $x^{u}$. Similarily,  we have $x^{o}\sim \mathcal{O}$ where $x^{o}\in \mathbf{O}$. We use  $P_U(x)$ and $P_O(x)$ to denote the probability density function for $\mathcal{U}$ and $\mathcal{O}$, respectively. Then the information loss for the spike firing process can be described as 
\begin{equation}\label{kl}
	\mathcal{L}_{KL}(\mathcal{U}||\mathcal{O})=\int_{-\infty}^{\infty}P_U(x^u) {\rm log} \frac{P_U(x^u) }{P_O(x^o)}dx.
\end{equation}
Since output spike $x^{o}$ is discrete, we can update it as 
\begin{equation}\label{kld}
	\mathcal{L}_{KL}(\mathcal{U}||\mathcal{O})=\int_{o-\epsilon}^{o+\epsilon}P_U(x^u) \log \frac{P_U(x^u)}{P_O(x^o)}dx,
\end{equation}
where $\epsilon$ is a small constant, $o$ is the spike value, $P_O(x^o)$ can be seen as a very large constant in this situation, and $\int_{o-\epsilon}^{o+\epsilon}P_O(x^o)dx=1$. With RMP-Loss, it will become
\begin{equation}\label{kl2}
	\mathcal{L}_{KL}(\hat{\mathcal{U}}||\mathcal{O})=\int_{o-\epsilon}^{o+\epsilon}P_{\hat{U}}(x^{\hat{u}}) \log \frac{P_{\hat{U}}(x^{\hat{u}})}{P_O(x^{o})}dx,
\end{equation}
where $x^{\hat{u}} \in\hat{\mathcal{U}}$ and $\hat{\mathcal{U}}$ is the new distribution of $x^{\hat{u}}$ adjusted by RMP-Loss.
Here, we have the following propositions.

\noindent \textbf{Proposition 1 } \textit{$\frac{d\mathcal{L}_{KL}}{d P_U(x^u)}<0$, \textit{i.e.}, $\mathcal{L}_{KL}$ $\downarrow$ as $P_U(x^u)$ $\uparrow$.}

\noindent \textbf{proof:} 
When a membrane potential $u$ is below the firing threshold, it will be pushed to 0 by the RMP-Loss, considering its effect, and otherwise, 1. 
Hence the firing rate of the neuron can be seen as the same, whether using the RMP-Loss or not, \textit{i.e.}, $P_O(x^o)$ keep the same for $\hat {\mathbf U}$ and $\mathbf U$. 
	\begin{align}
		\frac{d\mathcal{L}_{KL}}{d P_U(x^u)} &= \frac{d \int_{o-\epsilon}^{o+\epsilon}P_U(x^u) \log \frac{P_U(x^u)}{P_O(x^o)}dx}{d P_U(x^u)} \\
		&= \int_{o-\epsilon}^{o+\epsilon}(\log \frac{P_U(x^u)}{P_O(x^o)}+\frac{1}{\rm {ln2}} )dx.
	\end{align}
Since $P_O(x^o)$ is much larger than $ P_U(x^u)$, we can get $\int_{o-\epsilon}^{o+\epsilon}(\log \frac{P_U(x^u)}{P_O(x^o)}+\frac{1}{\rm {ln2}} )dx < 0$, then $\frac{d\mathcal{L}_{KL}}{d P_U(x^u)}<0$. $\blacksquare$

\noindent \textbf{Proposition 2 } \textit{ $P_{\hat{U}}(x^{\hat{u}}) > P_U(x^u)|x=o$.}

\noindent \textbf{proof:} 
We assume that there are $n^u_{\epsilon}$ samples in the interval $(o-\epsilon, o+\epsilon)$ for $\mathcal{U}$, where $\epsilon$ is a small constant. Then we can get $P_U(x^u) \approx \frac{n^u_{\epsilon}}{2\epsilon}$.
And we can simply see the RMP-Loss as a function to push the vanilla $\mathbf U$ to $ \hat {\mathbf U}$, which is closer to $\mathbf O$, based on its effect. 
Therefore, we can assume that these samples will be gathered to a new interval $(o-{\epsilon}_l, o+{\epsilon}_r)$, where ${\epsilon}_l,{\epsilon}_r < {\epsilon}$. Then we can get $P_{\hat{U}(x^{\hat{u}})} \approx \frac{n^u_{\epsilon}}{{\epsilon}_l + {\epsilon}_r}$.
Thereby we can have $P_{\hat{U}(x^{\hat{u}})} > P_U(x^u)|_{x=o}$. $\blacksquare$

Along with \textbf{Proposition 1} and \textbf{Proposition 2}, we can conclude that $\mathcal{L}_{KL}(\hat{U}||O)<\mathcal{L}_{KL}({U}||O)$, \textit{i.e.}, our method with RMP-Loss enjoys lesser information loss.

\section{Experiment}

In this section, we first conducted extensive ablation studies to compare the SNNs with RMP-Loss and their vanilla counterparts to verify the effectiveness of the method. Next, we fully compared our method with other state-of-the-art (SoTA) methods. Finally, some further experimental visualizations are provided to understand the RMP-Loss. 

\begin{table*}[h]
	\centering	
	\caption{Ablation Study for RMP-Loss}	
	\label{tab:ablation}
	 \setlength{\tabcolsep}{2.5mm}{
	\begin{tabular}{lllccc}	
		\toprule
		Dataset & Architecture & Type & Timestep & Accuracy & Quantization error \\	
		\toprule
		\multirow{6}{*}{CIFAR-10} & \multirow{2}{*}{ResNet20} & Without RMP-Loss & 4 & 91.26\% & 0.186  \\	

		                          &                           & With RMP-Loss & 4 & \textbf{91.89\%}  & 0.121 \\
		\cline{2-6}
		                          &\multirow{2}{*}{ResNet19}  & Without RMP-Loss & 4 & 95.04\%  & 0.128 \\	

		                          &                           & With RMP-Loss & 4 & \textbf{95.51\%}  & 0.104 \\
		\cline{2-6}
		                          &\multirow{2}{*}{VGG16}  & Without RMP-Loss & 4 & 92.80\% & 0.174  \\	
		                          &                           & With RMP-Loss & 4 & \textbf{93.33\%} & 0.135  \\		                          
		\bottomrule			   		         	            			         	
	\end{tabular}
	}
\end{table*}

\subsection{Datasets and Settings}

\textbf{Datasets.} We conducted experiments on four benchmark datasets: CIFAR-10 (100)~\cite{CIFAR-10}, CIFAR10-DVS~\cite{2017CIFAR10}, and ImageNet (ILSVRC12)~\cite{2009ImageNet}. The CIFAR-10 (100) dataset includes 60,000 images in 10 (100) classes with $32\times 32$ pixels. The numbers of training images and test images are 50,000 and 10,000. The CIFAR10-DVS dataset is the neuromorphic version of the CIFAR-10 dataset. It is composed of 10,000 images in 10 classes. We split the dataset into 9000 training images and 1000 test images similar to~\cite{2018Direct}. ImageNet dataset consists of 1,250,000 training images and 50,000 test images.

\textbf{Preprocessing.} We applied data normalization on all static datasets to make input images have $0$ mean and $1$ variance. Besides, we conducted random horizontal flipping and cropping on these datasets to avoid overfitting. For CIFAR, the AutoAugment~\cite{cubuk2019autoaugment} and  Cutout~\cite{devries2017improved} were also used for data augmentation as doing in~\cite{Guo_2022_CVPR,guoloss}. For the neuromorphic dataset, we resized the training image frames to $48\times 48$ as in~\cite{2020Going} and adopted random horizontal flip and random roll within $5$ pixels for augmentation. And the test images were merely resized to $48\times 48$ without any additional processing.

\textbf{Training setup.} For all the datasets, the firing threshold $V_{\rm th}$ was set as $0.5$. For static image datasets, the images were encoded to binary spike using the first layer of the SNN, as in recent works~\cite{2020DIET,2020Incorporating,2021Deep}. This is similar to rate coding. For the neuromorphic image dataset, we used the $0/1$ spike format directly. The neuron models in the output layer accumulated the incoming inputs without generating any spike as the output like in~\cite{2020DIET}. For CIFAR-10 (100) and CIFAR10-DVS datasets, we used the SGD optimizer with the momentum of $0.9$ and learning rate of $0.01$ with cosine decayed~\cite{2016SGDR} to $0$ as in~\cite{guo2022reducing}. All models were trained within $400$ epochs with the same batch size of $128$. For the ImageNet dataset, we adopted the SGD optimizer with a momentum of $0.9$ and a learning rate of $0.1$ with cosine decayed to $0$. All models are trained within $320$ epochs.

\subsection{Ablation Study for RMP-Loss}

\begin{figure}[h]
	\centering
	\includegraphics[width=0.50\textwidth]{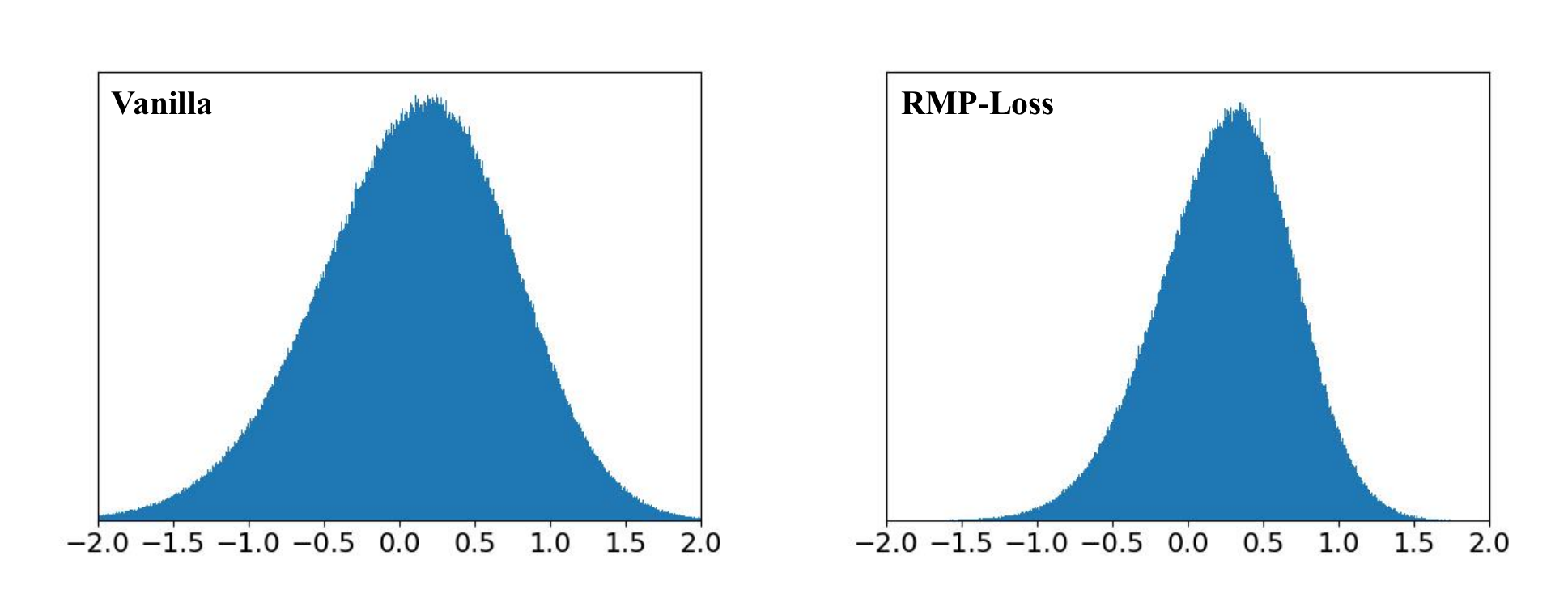}
	\caption{The effect of  RMP-Loss. The overall original membrane potential distribution (left) and the redistributed membrane potential distribution by  RMP-Loss (right) of the first layer of the second block in ResNet20 on CIFAR-10 test sets.}
	\label{fig:ablation}
\end{figure}

We conducted a set of ablation experiments on CIFAR-10 using ResNet20, ResNet19, and VGG16 as backbones. The results are shown in Tab.~\ref{tab:ablation}. It can be seen that with the RMP-Loss, these SNNs can achieve higher accuracy than their vanilla counterparts. We also show the membrane potential distribution of the first layer of the second block in the ResNet20 with and without RMP-Loss on the test set of CIFAR-10 in Fig.~\ref{fig:ablation}. It can be seen that the models trained with RMP-Loss can shrink the membrane potential distribution range which enjoys less quantization error. 

\begin{table*}[t]
	\centering	
	\caption{Comparison with SoTA methods on CIFAR-10/100.}	
	\label{tab:Comparisoncifar}	
	\begin{tabular}{lllccc}	
		\toprule
		Dataset & Method & Type & Architecture & Timestep & Accuracy \\	
		\toprule
		\multirow{28}{*}{CIFAR-10}	
		& SpikeNorm~\cite{2019Going} & ANN2SNN & VGG16 & 2500 & 91.55\%   \\	
		& Hybrid-Train~\cite{2020Enabling} & Hybrid training & VGG16 & 200 & 92.02\%   \\
		& Spike-Thrift~\cite{Kundu_2021_WACV} & Hybrid training & VGG16 & 100 & 91.29\%   \\
		& Spike-basedBP~\cite{2019Enabling} & SNN training & ResNet11 & 100 & 90.95\%   \\
		& STBP~\cite{2018Direct} & SNN training & CIFARNet & 12 & 90.53\%   \\  	
		& TSSL-BP~\cite{2020Temporal} & SNN training & CIFARNet & 5 & 91.41\%   \\ 
		& PLIF~\cite{2020Incorporating} & SNN training & PLIFNet & 8 & 93.50\%   \\
  	& DSR~\cite{meng2023training} & SNN training & ResNet18 & 20 & 95.40\%   \\
        & Joint A-SNN~\cite{guo2023joint} & SNN training & ResNet18 & 4 & 95.45\%   \\ 
        
		\cline{2-6}
		& \multirow{4}{*}{Diet-SNN~\cite{2020DIET}} & \multirow{4}{*}{SNN training} & \multirow{2}{*}{VGG16} & 5 & 92.70\%   \\ 
		& &  &  & 10 & 93.44\%   \\ 
		\cline{4-6}
		&  &  & \multirow{2}{*}{ResNet20} & 5 & 91.78\%   \\ 
		&  &  &  & 10 & 92.54\%   \\   
		\cline{2-6}
		& \multirow{3}{*}{RecDis-SNN~\cite{Guo_2022_CVPR}} & \multirow{3}{*}{SNN training} & \multirow{3}{*}{ResNet19} 
		& 2 & 93.64\%   \\
		&  &  &											                                  & 4 & 95.53\%   \\
		&  &  &											                                   & 6 & 95.55\%   \\
		\cline{2-6}
		& \multirow{3}{*}{STBP-tdBN~\cite{2020Going}} & \multirow{3}{*}{SNN training} & \multirow{3}{*}{ResNet19} 
		& 2 & 92.34\%   \\
		&  &  &											                                  & 4 & 92.92\%   \\
		&  &  &											                                   & 6 & 93.16\%   \\
		\cline{2-6}
		& \multirow{3}{*}{TET~\cite{deng2022temporal}} & \multirow{3}{*}{SNN training} & \multirow{3}{*}{ResNet19} 
		& 2 & 94.16\%   \\
		&  &  &											                                  & 4 & 94.44\%   \\
		&  &  &											                                   & 6 & 94.50\%   \\
		\cline{2-6}
		& \multirow{3}{*}{InfLoR-SNN~\cite{guo2022reducing}} & \multirow{3}{*}{SNN training} & \multirow{3}{*}{ResNet19} 
		& 2 & 94.44\%   \\
		&  &  &											                                  & 4 & 96.27\%   \\
		&  &  &											                                   & 6 & 96.49\%   \\
		\cline{2-6}
		& \multirow{7}{*}{\textbf{RMP-Loss}} & \multirow{7}{*}{SNN training} & \multirow{3}{*}{ResNet19} 
		& 2 & \textbf{95.31\%}$\pm 0.07$  \\
		&  &  &											                                  & 4 & \textbf{95.51\%}$\pm 0.08$  \\
		&  &  &											                                   & 6 & \textbf{96.10\%}$\pm 0.08$   \\	
		\cline{4-6}	
		&  &  & \multirow{2}{*}{ResNet20} 		                                          & 4 & \textbf{91.89\%}$\pm 0.05$   \\
		&  &  &											                                  & 6 & \textbf{92.55\%}$\pm 0.06$   \\
		\cline{4-6}		
		&  &  &	\multirow{2}{*}{VGG16}                                                   & 4 & \textbf{93.33\%}$\pm 0.07$  \\
		&  &  &											                                 & 10 & \textbf{94.39\%}$\pm 0.08$   \\		
		\hline	
		
		\multirow{16}{*}{CIFAR-100}	
    	& DSR~\cite{meng2023training} & SNN training & ResNet18 & 20 & 78.50\%   \\
         & InfLoR-SNN~\cite{guo2022reducing} & SNN training & VGG16 & 5 & 71.56\%   \\
         & IM-Loss~\cite{guoloss} & SNN training & VGG16 & 5 & 70.18\%   \\ 
		\cline{2-6} 
		& \multirow{2}{*}{Diet-SNN~\cite{2020DIET}} & \multirow{2}{*}{SNN training} & ResNet20 & 5 & 64.07\%   \\ 
		&                                          &                                & VGG16 & 5 & 69.67\%   \\  
		\cline{2-6}
		& \multirow{2}{*}{Real Spike~\cite{guo2022real}} & \multirow{2}{*}{SNN training} & ResNet20 & 5 & 66.60\%   \\ 
		&                                          &                                & VGG16 & 5 & 70.62\%   \\  
		\cline{2-6}
		& \multirow{3}{*}{TET~\cite{deng2022temporal}} & \multirow{3}{*}{SNN training} & \multirow{3}{*}{ResNet19} 
		& 2 & 72.87\%   \\
		&  &  &											                                  & 4 & 74.47\%   \\
		&  &  &											                                   & 6 & 74.72\%   \\
		\cline{2-6}
		& \multirow{6}{*}{\textbf{RMP-Loss} } & \multirow{6}{*}{SNN training} & {ResNet20} 
		& 4 & \textbf{66.65\%}$\pm 0.10$   \\
		\cline{4-6}	
		&  &  &	\multirow{2}{*}{VGG16}                                                       & 4 & \textbf{72.55\%}$\pm 0.08$   \\
		&  &  &											                                     & 10 & \textbf{73.30\%}$\pm 0.11$   \\
		\cline{4-6}	
		&  &  &	\multirow{3}{*}{ResNet19}                                                    & 2 & \textbf{74.66\%}$\pm 0.12$  \\
		&  &  &											                                     & 4 & \textbf{78.28\%}$\pm 0.10$  \\
		&  &  &											                                     & 6 & \textbf{78.98\%}$\pm 0.08$   \\	
		\bottomrule				         	
	\end{tabular}	
\end{table*}

\subsection{Comparison with State-of-the-art Methods}

We evaluated the proposed method with the accuracy performance on various widely used static and neuromorphic datasets using spiking ResNet20~\cite{2020DIET,2019Going}, VGG16~\cite{2020DIET}, ResNet18~\cite{2021Deep},  ResNet19~\cite{2020Going}, and ResNet34~\cite{2021Deep}. The results with the mean accuracy and standard deviation of 3-trials are listed in Tab.~\ref{tab:Comparisoncifar}.

\textbf{CIFAR-10.}
For CIFAR-10,  we tested three different network architectures. The performance of our method is shown in Tab.~\ref{tab:Comparisoncifar}. It can be seen that our method achieves the best results in most of these cases. In special, using ResNet19~\cite{2020Going}, the RMP-Loss achieves 96.10\% averaged top-1 accuracy, which improves 1.60\% absolute accuracy compared with the existing state-of-the-art TET~\cite{deng2022temporal}. Though InfLoR-SNN~\cite{guo2022reducing} is slightly better than our method for 4 and 6 timesteps, it will decrease the biological plausibility of the SNN and increase the inference burden since a complex transform function is added in its spiking neuron.
On ResNet20, our method can achieve higher accuracy with only 6 timesteps, while Diet-SNN~\cite{2020DIET} with 10 timesteps. On VGG16, the RMP-Loss also shows obvious advantages. 

\textbf{CIFAR-100.}
For CIFAR-100, we also experimented with these three different network structures. For all these configurations, our method achieves consistent and significant accuracy over prior work. ResNet20-based and VGG16-based RMP-Loss achieve 66.65\% and 72.55\% top-1 accuracy with only 4 timesteps, which outperform their Diet-SNN counterparts with 2.58\% and 2.88\% higher accuracy and Real spike counterparts with 0.05\% and 1.93\% respectively but with fewer timesteps. Noteworthy, our method significantly surpasses TET with 4.26\% higher accuracy on ResNet19, which is not easy to achieve in the SNN field. Overall, our results on CIFAR-100 show that, when applied to a more complex dataset, the RMP-Loss achieves an even more favorable performance compared to competing methods.

\begin{table*}[htbp]
	\centering	
	\caption{Comparison with SoTA methods on CIFAR10-DVS.}	
	\label{tab:Comparisondvs}	
	\begin{tabular}{lllccc}	
		\toprule
		Dataset & Method & Type & Architecture & Timestep & Accuracy \\	
		\toprule
		\multirow{9}{*}{CIFAR10-DVS}	
		& Rollout~\cite{2020Efficient} & Rollout & DenseNet & 10 & 66.80\%   \\	
		& STBP-tdBN~\cite{2020Going} & SNN training & ResNet19 & 10 & 67.80\%   \\ 
		& LIAF~\cite{Wu_2021} & Conv3D & LIAF-Net & 10 & 71.70\%   \\ 
		& LIAF~\cite{Wu_2021} & LIAF & LIAF-Net & 10 & 70.40\%   \\
       & RecDis-SNN~\cite{Guo_2022_CVPR} & SNN training & ResNet19 & 10 & 72.42\%   \\
        \cline{2-6}
		& \multirow{2}{*}{InfLoR-SNN~\cite{guo2022reducing}} & \multirow{2}{*}{SNN training} & {ResNet19} 
		& 10 & 75.50\%   \\
		&  &  &											                 {ResNet20} & 10 & 75.10\%   \\
		\cline{2-6}
		& \multirow{2}{*}{\textbf{RMP-Loss}} & \multirow{2}{*}{SNN training} & {ResNet19} 
		& 10 & \textbf{76.20\%}$\pm 0.20$   \\
		&  &  &											                 {ResNet20} & 10 & \textbf{75.60\%}$\pm 0.30$   \\
		\bottomrule				         	
	\end{tabular}	
\end{table*}

\begin{table*}[htbp]
	\centering	
	\caption{Comparison with SoTA methods on ImageNet.}	
	\label{tab:Comparisonimage}	
	\begin{tabular}{llcccc}	
		\toprule
		Method & Type & Architecture & Timestep & Spike form& Accuracy \\	
		\toprule
			
		Hybrid-Train~\cite{2020Enabling} & Hybrid training& ResNet34 & 250 &Binary & 61.48\%   \\  
		SpikeNorm~\cite{2019Going} & ANN2SNN & ResNet34 & 2500 &Binary& 69.96\%   \\		
		STBP-tdBN~\cite{2020Going} &  SNN training & ResNet34 & 6 &Binary& 63.72\%   \\ 
		TET~\cite{deng2022temporal} &  SNN training & ResNet34 & 6 &Binary& 64.79\%   \\
		\cline{1-6}
		\multirow{2}{*}{SEW ResNet~\cite{2021Deep}} & \multirow{2}{*}{SNN training} & {ResNet18} & 4 &Integer & \textbf{63.18\%}   \\
		 &  &											                             {ResNet34} & 4 & Integer & \textbf{67.04\%}   \\ 
		\cline{1-6}
		\multirow{2}{*}{Spiking ResNet~\cite{2021Deep}} & \multirow{2}{*}{SNN training} & {ResNet18} & 4 &Binary& {62.32\%}   \\
		 &  &											                             {ResNet34} & 4 &Binary& {61.86\%}   \\ 
		\cline{1-6}
		\multirow{2}{*}{\textbf{RMP-Loss} } & \multirow{2}{*}{SNN training} & {ResNet18} & 4 &Binary& 63.03\%$\pm 0.07$   \\
		 &  &											                             {ResNet34} & 4 &Binary& 65.17\%$\pm 0.07$   \\
		\bottomrule				         	
	\end{tabular}	
\end{table*}

\textbf{CIFAR10-DVS.}
We also verified our method on the popular neuromorphic dataset, CIFAR10-DVS. It can be seen that RMP-Loss also shows amazing performance. RMP-Loss outperforms STBP-tdBN by 12.39\%, RecDis-SNN by 3.78\%, and InfLoR-SNN by 0.70\% respectively with 76.20\% top-1 accuracy in 10 timesteps using ResNet19 as the backbone. With ResNet20 as the backbone, RMP-Loss can also achieve well-performed results.

\textbf{ImageNet.}
For ImageNet, we conducted experiments with ResNet18 and ResNet34 as backbones. Our results are presented in Tab.~\ref{tab:Comparisonimage}. In these normal spiking structures, our method achieves the highest accuracy among these SoTA prior works. However, our method is slightly worse than SEW ResNet~\cite{2021Deep}. This is because that SEW ResNet uses an atypical architecture, which abandons binary spike forms but will output arbitrary integers to transmit information, thus enjoying higher accuracy but the efficiency from multiplication free of SNNs will be lost. Hence, it would be limited in some applications while our method not.

\subsection{Visualization}

\begin{figure}[h]
	\centering
	\includegraphics[width=0.50\textwidth]{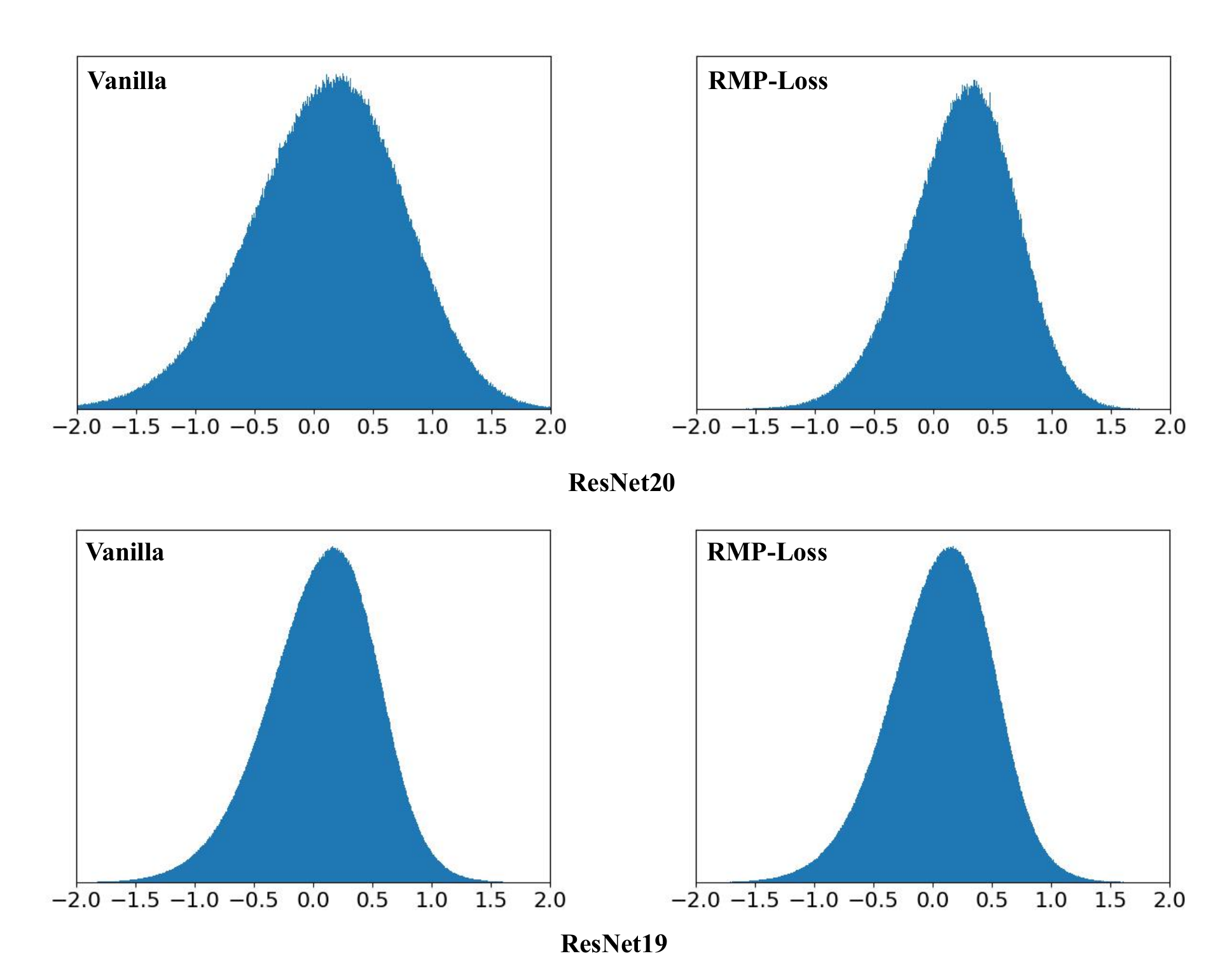} 
	\caption{The original membrane potential distribution (left) and the redistributed membrane potential distribution by  RMP-Loss (right) of the first layer of the second block in ResNet20/19 on CIFAR-10 test sets.}
	\label{distribution}
\end{figure}
Furthermore, we also provide some experimental visualizations to show the regularizing effects of RMP-Loss. Fig.~\ref{distribution} shows the membrane potential distribution of the first layer of the second block in the ResNet20/19 with and without RMP-Loss on the test set of CIFAR-10. It can be seen that the SNN models trained with RMP-Loss can shrink the membrane potential distribution range which enjoys less quantization error. On the other hand, comparing the membrane potential distribution difference between ResNet20 and ResNet19, it can be found that the membrane potential distribution of ResNet19 is thinner than that of ResNet20 too. Considering that ResNet19 can achieve higher accuracy than ResNet20, it also shows that reducing quantization error is effective to improve the accuracy of SNN models and our route to improve the SNN accuracy by reducing the quantization error is reasonable.

\section{Conclusion}

This paper aims at addressing the information loss problem caused by the $0/1$ spike quantization of SNNs. We introduce RMP-Loss to adjust the membrane potential to reduce the quantization error. Different from other methods that reduce the quantization error indirectly or will induce more parameters, RMP-Loss focuses on handling this problem directly and will introduce no extra parameters in the inference phase. We show that our method outperforms SoTA methods on both static and neuromorphic datasets.

\section*{Acknowledgment}
This work is supported by grants from the National Natural Science Foundation of China under
contracts No.12202412 and No.12202413.

{\small
\bibliographystyle{ieee_fullname}
\bibliography{egbib}
}

\end{document}